\documentclass[11pt]{article}

\usepackage[final]{acl}
\usepackage{times}
\usepackage{latexsym}
\usepackage[T1]{fontenc}
\usepackage[utf8]{inputenc}
\usepackage{microtype}
\usepackage{inconsolata}
\usepackage{amsmath}
\usepackage{graphicx}
\usepackage{subcaption}
\usepackage{booktabs}
\usepackage[nointegrals]{wasysym}

\newcommand{\yes}{\CIRCLE}
\newcommand{\prt}{\LEFTcircle}
\newcommand{\nope}{\Circle}
\newcommand{\nsc}{\textendash}

\DeclareRobustCommand{\VL}{Dr.~Claw}

\title{\VL{}: An AI Scientist Workspace for Vibe Research}

\author{
  \textbf{Dingjie Song$^{1}$, Hanrong Zhang$^{2}$, Dawei Liu$^{3}$, Yixin Liu$^{1}$, Zongxia Li$^{4}$,} \\
  \textbf{Zhengqing Yuan$^{5}$, Siqi Zhang$^{1}$, Henry Peng Zou$^{2}$, Zhiling Yan$^{1}$, Yuxuan Zhang$^{6}$,} \\
  \textbf{Yanfang Ye$^{5}$, Philip S. Yu$^{2}$, Lichao Sun$^{1}$} \\
  $^{1}$Lehigh University \quad
  $^{2}$University of Illinois Chicago \quad
  $^{3}$University of Pennsylvania \\
  $^{4}$University of Maryland \quad
  $^{5}$University of Notre Dame \quad
  $^{6}$University of British Columbia \\
}

\newcommand{\oeBarePool}{0.873}\newcommand{\oeDrPool}{0.952}
\newcommand{\oeDeltaPool}{+0.079}
\newcommand{\oeCIPool}{[-0.00,\,+0.14]}
\newcommand{\oeSubgroupBare}{0.33}\newcommand{\oeSubgroupDr}{1.00}
\newcommand{\oeLimitsBare}{0.33}\newcommand{\oeLimitsDr}{1.00}
\newcommand{\oeRelworkBare}{0.00}\newcommand{\oeRelworkDr}{0.67}
\newcommand{\oeGraphTasks}{17}\newcommand{\oeSkillReads}{12}
\newcommand{\oeAuTaskGraph}{14}\newcommand{\oeAuExecTrace}{14}
\newcommand{\oeAuDecLog}{10}
\newcommand{\oeAuXrefBarePct}{62\%}\newcommand{\oeAuXrefDrPct}{100\%}

\begin{document}
\maketitle

\begin{abstract}
Command-line coding agents (e.g., Claude Code, Gemini CLI) can already read and write files and sustain long sessions, yet end-to-end research still fragments across chat tools, IDEs, terminals, and writing environments, and the decisions that make it auditable are rarely preserved.
We present \textbf{\VL{}}, an open-source workspace that wraps existing coding-agent executors in a \emph{controllable} and \emph{auditable} human-in-the-loop workflow rather than introducing another autonomous agent. Persistent state objects, a reusable skill library, and multi-executor coordination link human decisions to AI execution, turning planning, execution, and writing into one traceable, recoverable loop.
We demonstrate \VL{} through an interactive three-view scenario and a failure-recovery walkthrough, and evaluate it against a bare command-line agent sharing the same backend executor, so the comparison contrasts the whole orchestration layer (task graph, state objects, and skill library) with the agent it wraps. Holding the executor fixed, \VL{} scores higher on research completeness while persisting an auditable, recoverable process trail. Demo access: repository \url{https://github.com/OpenLAIR/dr-claw}, released under AGPL-3.0 with GPL-3.0 upstream components.
\end{abstract}

\section{Introduction}

Large foundation models and agentic tools have improved the five core AI research operations---literature review, idea generation, code implementation, results analysis, and drafting \citep{openai2023gpt4,Brown2020LanguageMA,Yao2022ReActSR,Schick2023Toolformer,lu2024aiscientist,yu2025tinyscientist,yang2025researstudio,schmidgall2025agentlaboratory,baek2025researchagent}. Command-line coding agents such as Claude Code and Gemini CLI push this further, living in the terminal, reading and writing project files, and sustaining context across long sessions \citep{chen2021evaluating,barke2022grounded,moradi2022github,Jimenez2024SWEbench,yang2024sweagent}. Yet these agents optimize execution, not \emph{control}: the plan, intermediate decisions, and artifacts that make a research process reviewable are scattered or lost, and the human has few explicit takeover points.

\begin{figure}[!t]
  \centering
  \includegraphics[width=\columnwidth]{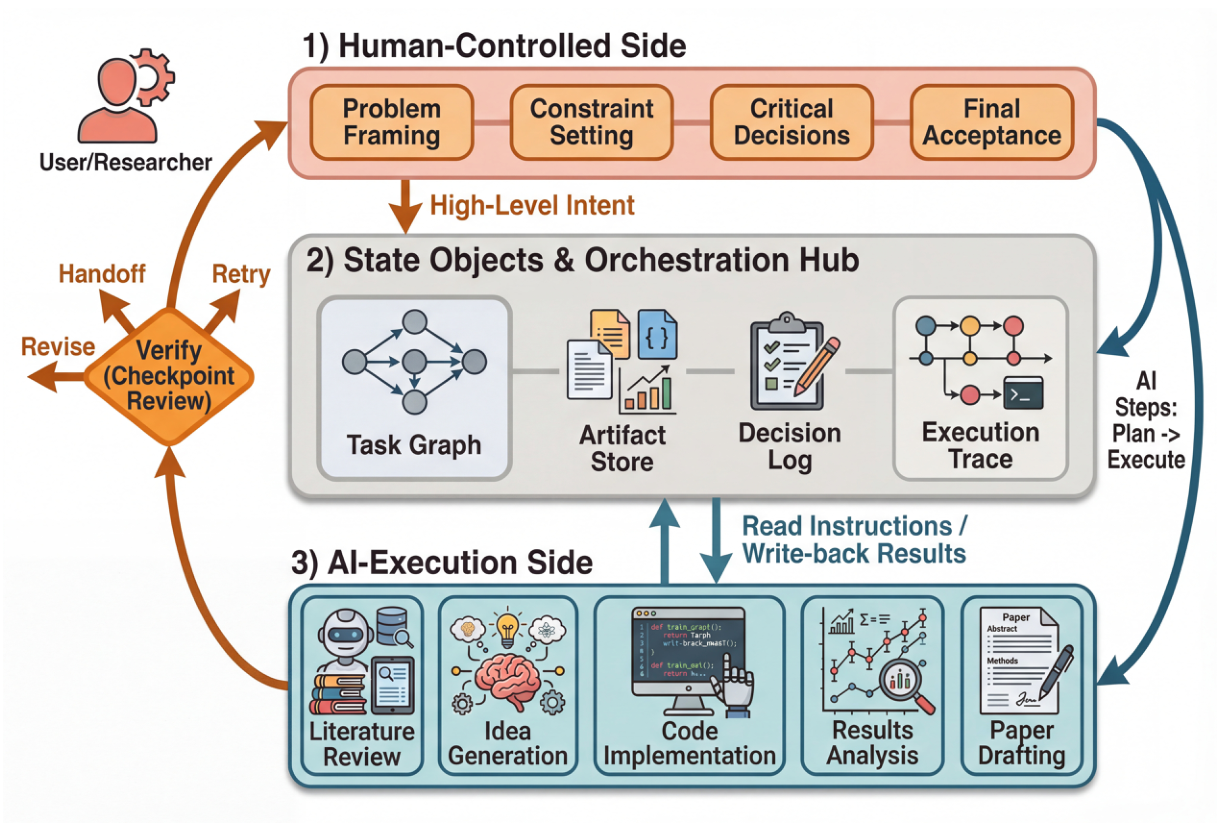}
  \vspace{-2mm}
  \vspace{-2mm}
  \vspace{-2mm}
  \vspace{-2mm}
  \caption{Control split in \VL{}: the Human-Controlled side sets goals, constraints, and acceptance decisions; the AI-Execution side runs the five core AI operations (Plan $\rightarrow$ Execute), connected through four state objects and checkpoint feedback (Verify/Revise/Retry/Handoff).}
  \label{fig:layered-framework}
  \vspace{-2mm}
  \vspace{-2mm}
\end{figure}

The bottleneck is now full-process orchestration rather than isolated capability: researchers still switch across tools for decomposition, scheduling, tracking, validation, and writing, which weakens reproducibility and delivery reliability. HCI evidence consistently shows that collaboration cost is dominated by verification and context maintenance, and that process visibility and interruptible control are critical \citep{gu2024analystsverify,kazemitabaar2024steeringverification,xie2024waitgpt,codeaid2024,flores2025impact}; existing demos improve usability but leave cross-stage state continuity and artifact closed-loop management limited \citep{dibia2024autogenstudio,cai2024lowcodellm}.

We target a mode of work we call \textbf{Vibe Research}: a controllable, human-in-the-loop paradigm in which a researcher states high-level goals and constraints in natural language, AI compiles them into an executable research loop and carries out the five core operations, and final acceptance rests on observable outputs, with humans governing direction and final decisions throughout. The paradigm is defined not by full autonomy but by an operational human--AI division of labor: AI handles high-throughput, parallelizable, templatable execution (retrieval, coding, running, summarizing, drafting), while humans own research direction, evaluation criteria, key trade-offs, and final acceptance. Unlike end-to-end autonomous research agents \citep{lu2024aiscientist,yamada2025aiscientistv2,tang2025airesearcher,schmidgall2025agentlaboratory} or general multi-agent frameworks \citep{wu_autogen_2023,qian2024chatdev}, we emphasize sustained human takeover and research-centric artifact management (Section~2). Crucially, \VL{} does not introduce yet another executor: it \emph{wraps} an existing command-line coding agent, adding the state, control, and audit layer that such agents lack while reusing their execution capability.

We propose \textbf{\VL{}}, a one-stop workspace that unifies planning, execution, and writing into one controllable, traceable, recoverable, and auditable research loop (Figure~\ref{fig:layered-framework}). In each cycle, users provide goals, constraints, and acceptance criteria; the system decomposes tasks, executes actions, writes back artifacts, and supports revise/retry/handoff without losing process state. We evaluate this loop in Section~5, holding the backend executor fixed so that the comparison contrasts the orchestration layer as a whole with the bare executor, and complement it with a retrospective human study on efficiency, output quality, and integrated experience (Appendix~\ref{app:human-study}).
Our contributions are as follows:
\begin{itemize}
\item We formalize \emph{Vibe Research}, a controllable, human-in-the-loop research-orchestration paradigm that clarifies the boundary between AI execution and human decision responsibilities, distinguishing it from end-to-end autonomous approaches.
\item We implement this paradigm in \VL{}, with a task-graph-centric orchestration, a chat-driven planner, a modular skill library ($58$ stage-mapped skills across five research stages, $171$ in the deployed catalogue), and a multi-agent execution layer compatible with mainstream coding-agents.
\item We provide a controlled pilot evaluation and a scenario-based demonstration. Holding the executor fixed, \VL{} scores higher than the bare agent on completeness by closing its research-hygiene gaps (consistent across tasks, though not statistically powered at one run per task), while uniquely persisting an auditable, recoverable process trail; a retrospective study additionally associates the integrated workflow with gains in efficiency, quality, and usability over non-integrated ones.
\end{itemize}

\section{Related Work}

\begin{table*}[t]
  \centering
  \small
  \setlength{\tabcolsep}{4pt}
  \begin{tabular}{l ccccccc}
    \toprule
    & \multicolumn{2}{c}{\emph{Execution}} & \multicolumn{2}{c}{\emph{Orchestration}} & \multicolumn{2}{c}{\emph{Interaction}} & \\
    \cmidrule(lr){2-3} \cmidrule(lr){4-5} \cmidrule(lr){6-7}
    \textbf{System} &
    \textbf{Wraps CLI} & \textbf{Research} & \textbf{Built-in} & \textbf{In-place} & \textbf{Mid-run} & \textbf{End-user} & \textbf{Unattended} \\
    &
    \textbf{agent} & \textbf{skills} & \textbf{research} & \textbf{recovery} & \textbf{takeover} & \textbf{workspace} & \textbf{end-to-end} \\
    &
    & & \textbf{state} & & & & \\
    \midrule
    \multicolumn{8}{l}{\emph{End-to-end autonomous research systems}} \\
    \quad AI Scientist v1/v2& \prt  & \prt  & \yes  & \prt  & \nope & \nope & \yes \\
    \quad Agent Laboratory  & \nope & \prt  & \prt  & \prt  & \prt  & \nope & \yes \\
    \quad ResearchAgent     & \nope & \nope & \prt  & \nsc  & \nope & \nope & \prt \\
    \addlinespace
    \multicolumn{8}{l}{\emph{Agent authoring tools and orchestration runtimes}} \\
    \quad AutoGen Studio    & \nope & \nope & \nope & \nope & \prt  & \prt  & \yes \\
    \quad Flowise (archived)& \nope & \prt  & \nope & \yes  & \prt  & \yes  & \yes \\
    \quad LangGraph         & \nope & \nope & \nope & \yes  & \prt  & \nope & \yes \\
    \addlinespace
    \multicolumn{8}{l}{\emph{Intervenable research agents}} \\
    \quad TinyScientist     & \yes  & \prt  & \prt  & \prt  & \nope & \yes  & \yes \\
    \quad ResearStudio      & \nope & \prt  & \prt  & \yes  & \yes  & \yes  & \yes \\
    \quad IRIS              & \nope & \nope & \prt  & \prt  & \prt  & \yes  & \prt \\
    \midrule
    \textbf{\VL{}} (ours)   & \yes  & \yes  & \yes  & \yes  & \yes  & \yes  & \yes \\
    \bottomrule
  \end{tabular}
  \caption{\textbf{Design-space comparison.} \yes{} supported, \prt{} partial, \nope{} absent, \nsc{} out of scope by design.
  \emph{Built-in research state}: research-artifact objects (task graph, artifact store, decision log, execution trace) shipped with the system rather than a schema the developer declares.
  \emph{Mid-run takeover}: intervention at arbitrary points during a run, not only at stage boundaries or developer-placed pause nodes.
  Marks are assessed from published papers, official documentation, and public repositories as of August~2026, not from runs of these systems.
  Each dimension is realized, fully or in part, by prior systems; \VL{} differs by combining them.}
  \label{tab:design-space}
\end{table*}

\subsection{Research Agents and End-to-End Automation}

End-to-end systems automate the path from ideas to papers with minimal human intervention \citep{lu2024aiscientist,yamada2025aiscientistv2,schmidgall2025agentlaboratory,baek2025researchagent}, but rarely prioritize controllability in sustained collaboration within real local engineering environments. A parallel line lowers the barrier to building or steering agents: no-/low-code authoring \citep{dibia2024autogenstudio,cai2024lowcodellm,flowise2026} and general orchestration runtimes such as LangGraph \citep{langgraph2026}, which give developers durable checkpointing, human-in-the-loop interrupts, and replay over a state schema they declare themselves; TinyScientist \citep{yu2025tinyscientist}, ResearStudio \citep{yang2025researstudio}, and IRIS \citep{garikaparthi2025iris} add intervenable agents. \VL{} differs along three axes taken together: it (i) \emph{wraps} an existing command-line coding agent rather than introducing a new executor; (ii) makes the research process a first-class object through four persistent state objects (Task Graph, Artifact Store, Decision Log, Execution Trace); and (iii) sustains human takeover across the full ideation$\rightarrow$experiment$\rightarrow$publication arc (Table~\ref{tab:design-space}). Each axis has precedent taken alone---TinyScientist also delegates to an external coding agent, AI Scientist-v2 persists a structured experiment tree, and ResearStudio allows a pause at any moment rather than at checkpoints, where it is stronger than \VL{}---but prior work optimizes end-to-end autonomy, agent \emph{construction}, or steering within a single stage, whereas \VL{} optimizes \emph{controllability and auditability of an existing agent} over a long-horizon workflow.

\subsection{Human--AI Collaboration and Context-Switching Costs}

Recent HCI research has shifted from code-generation utility to process controllability and verification burden. In data analysis and knowledge work, the key bottlenecks are cross-step interpretation, validation, and correction rather than one-off generation quality, and interactive decomposition and process visualization improve monitoring and intervention \citep{gu2024analystsverify,kazemitabaar2024steeringverification,xie2024waitgpt}; in programming settings, CHI evidence shows the added reading, confirming, and revising of assistant outputs harms fluency, cognitive load, and confidence \citep{codeaid2024,flores2025impact}. These converge on a design requirement---clear supervision affordances, interpretable intermediate states, and timely takeover---that aligns with \VL{}'s goal of reducing context-switching costs. Accordingly, we evaluate the systemic effect of workflow integration on efficiency, quality, and usability rather than intrinsic model gains.

\section{System Overview}

\VL{} is designed around one central question: how to let researchers complete problem definition, experiment progression, and paper production in a continuous workflow rather than switching among isolated tools. It models research as an explicitly traceable workflow that organizes human decisions with AI execution into a stable collaboration structure. Additional implementation, failure-recovery, and reproducibility details are in Appendix~\ref{app:impl-details}, Appendix~\ref{app:failure-recovery}, and Appendix~\ref{app:repro-details}.

\begin{figure}[t]
  \centering
  \includegraphics[width=\columnwidth]{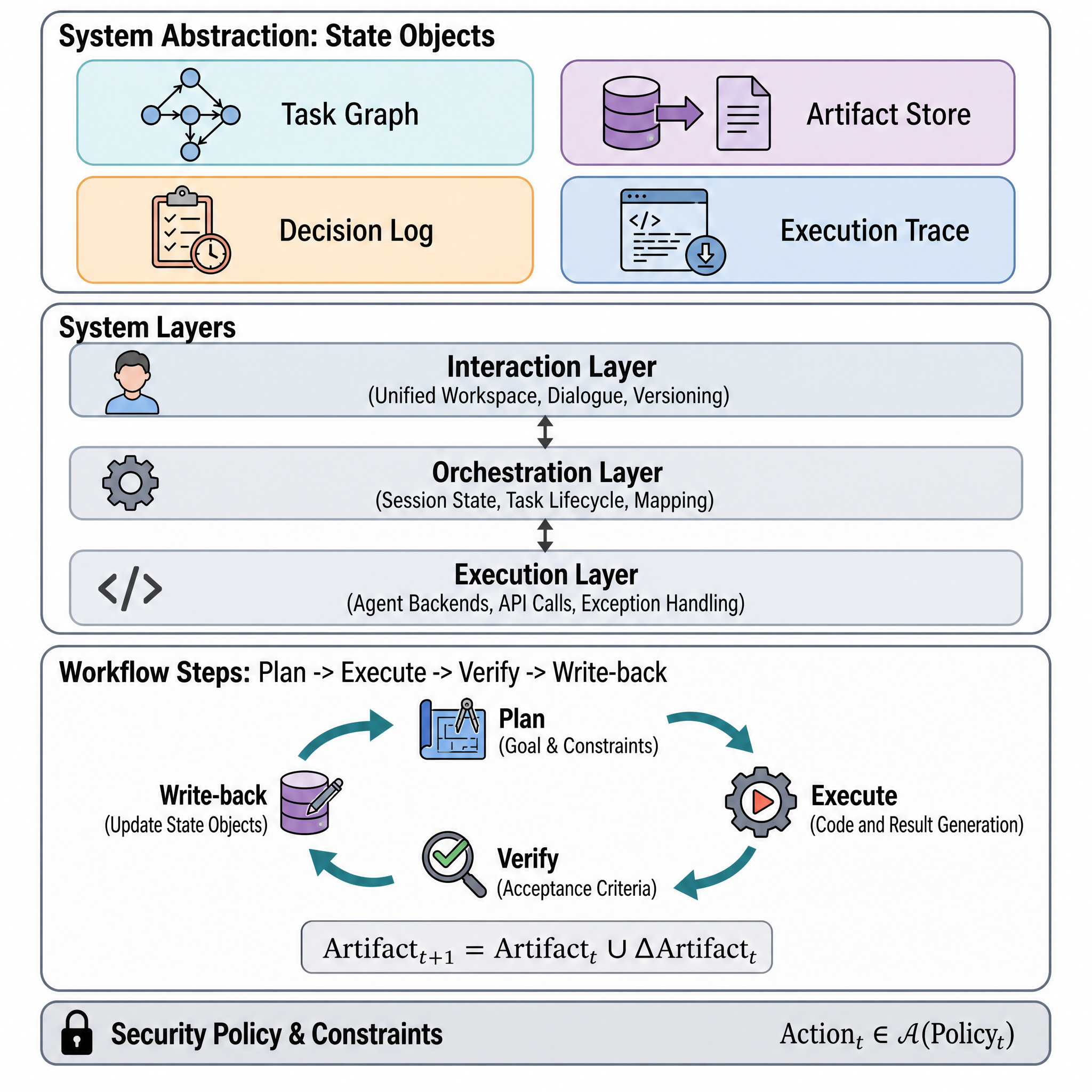}
  \caption{\textbf{Overall system view of \VL{}.} \textbf{Top}: four state objects (Task Graph, Artifact Store, Decision Log, Execution Trace). \textbf{Middle}: three system layers (Interaction, Orchestration, Execution). \textbf{Bottom}: iterative workflow steps (Plan--Execute--Verify--Write-back). \textbf{Lowest}: safety policy constraint ($\mathrm{Action}_t \in \mathcal{A}(\mathrm{Policy}_t)$).}
  \label{fig:system-overview}
\end{figure}

\subsection{Design Goals and System Abstraction}

\VL{} orchestrates four state objects: \texttt{Task Graph}, \texttt{Artifact Store}, \texttt{Decision Log}, and \texttt{Execution Trace}. Together they convert interaction history into a reviewable process supporting iterative orchestration rather than one-shot generation: users provide high-level goals/constraints, and the system maps them to executable tasks with continuous inspection and takeover. One task interaction is a state transition:
\begin{equation}
\mathrm{State}_{t+1}=f(\mathrm{State}_t,\mathrm{Action}_t,\mathrm{Obs}_t),
\end{equation}
where $\mathrm{State}_t$ denotes the workflow state at time $t$ (jointly formed by Task Graph, Artifact Store, Decision Log, and Execution Trace), $\mathrm{Action}_t$ is a system- or user-triggered action, and $\mathrm{Obs}_t$ is the observed feedback. This formulation highlights that \VL{} optimizes iterative state updates rather than single responses.

\subsection{Three-Layer System Architecture}

\VL{} uses three collaborative layers:
\begin{itemize}
\item \textbf{Interaction}: unified workspace for chat, task views, files, and version operations.
\item \textbf{Orchestration}: state and lifecycle management from high-level intent to stage tasks.
\item \textbf{Execution}: backend invocation, result return, and exception handling across heterogeneous executors.
\end{itemize}
This design enables backend switching without changing workflow semantics while preserving unified state and audit views.

\subsection{Workflow-Centric Interaction Loop}

\looseness=-1 Given a research idea, the system generates a structured brief and dependency-aware task plan, then runs workflow steps of Plan--Execute--Verify--Write-back. Execution outputs are written into \texttt{Artifact Store}, task states are updated in \texttt{Task Graph}, and all interventions are retained in \texttt{Decision Log}/\texttt{Execution Trace}. This workflow supports long-horizon iteration with explicit human checkpoints.

\subsection{Skill-Based Capabilities and Multi-Executor Coordination}

\VL{} provides a reusable skill library ($58$ stage-mapped skills across five research stages---survey, ideation, experiment, publication, promotion---and $171$ skills in the deployed catalogue) covering ideation, literature processing, experimentation, analysis, and writing. Each skill is a directory with a \texttt{SKILL.md} manifest whose YAML frontmatter declares a name and description, and commonly a version, license, allowed tools, and stage/domain tags. Skills are versioned and schema-checked before activation, and reach a task by three routes: a stage-skill map resolves the task's stage and type to a set of suggested skills, which are attached to the task node and injected into its next-action prompt; keyword detection over user instructions and task text can auto-load a skill; and users may invoke any catalogue skill manually from the Skills dashboard. Authoring, validation, and transfer are detailed in Appendix~\ref{app:impl-details}. \VL{} also coordinates multiple executors in one project context, so users switch execution strategy by task type, and on failure or constraint violation can retry, revise, or take over without breaking global state. Formalization of artifact updates is in Appendix~\ref{app:system-details}.

\begin{figure*}[t]
  \centering
  \includegraphics[width=1\textwidth]{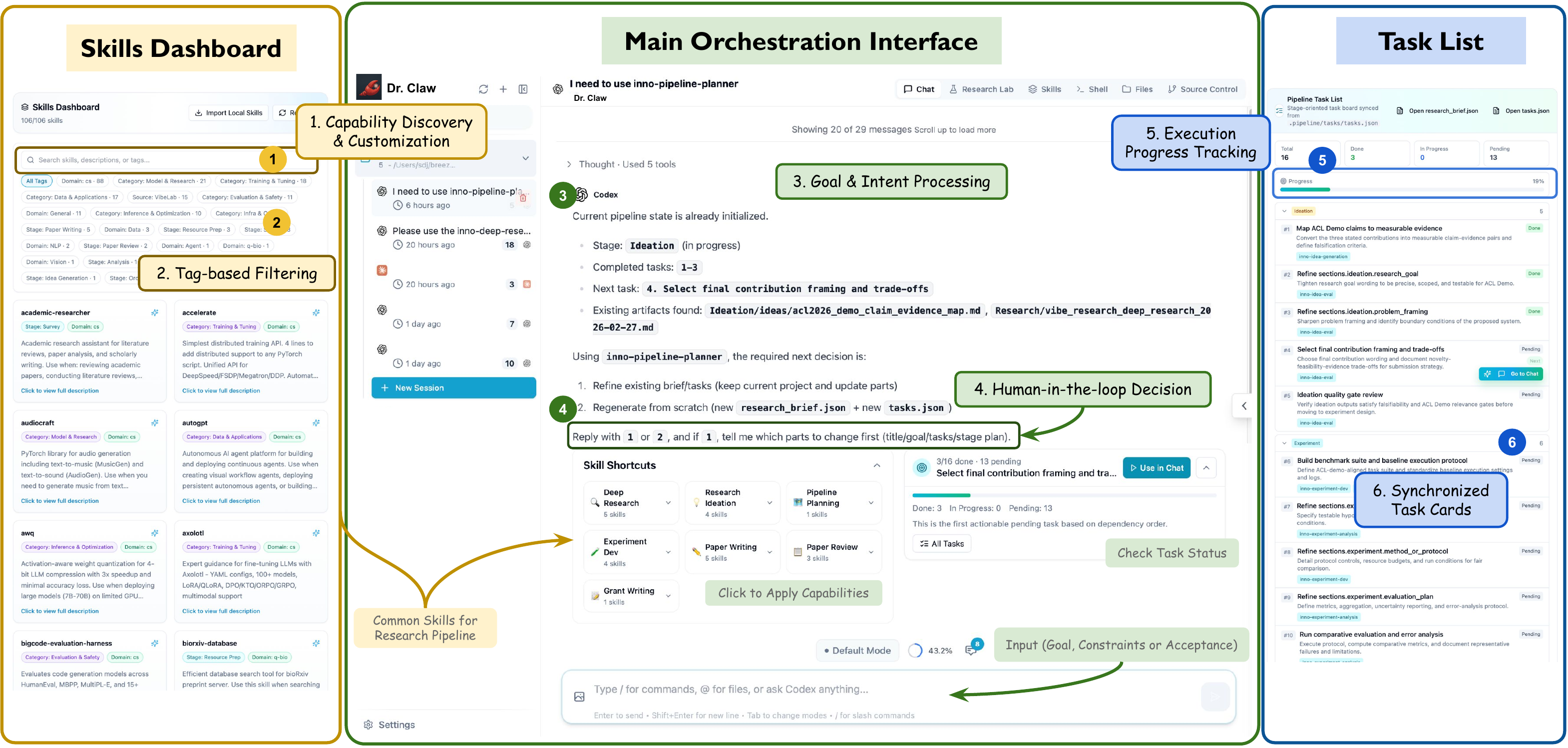}
  \vspace{-1mm}
  \vspace{-2mm}
  \vspace{-2mm}
  \vspace{-2mm}
  \caption{\textbf{Three-view scenario diagram}. \textbf{Left}: Skills Dashboard for capability discovery and filtering. \textbf{Center}: main orchestration interface for planning, execution feedback, and human approval. \textbf{Right}: Task List for progress tracking and next-step selection.}
  \label{fig:demo-planning}
  \vspace{-4mm}
\end{figure*}

\subsection{Safety and Controllability Mechanisms}

Given risks from external calls and code execution, \VL{} treats permission management as a first-class mechanism, supporting fine-grained tool/command policies that distinguish secure defaults from trusted extended settings. Actions are allowed only if they belong to the action set defined by the current policy:
\begin{equation}
\mathrm{Action}_t \in \mathcal{A}(\mathrm{Policy}_t),
\end{equation}
where $\mathrm{Policy}_t$ is the permission configuration at time $t$ and $\mathcal{A}(\mathrm{Policy}_t)$ is the executable action space, guaranteeing consistency between execution capability and safety boundaries.

\section{Demo Scenario}

We use a research task to illustrate \VL{} under human-in-the-loop conditions---whether high-level research intent can be stably transformed into executable workflows while preserving controllability, recoverability, and auditability---focusing on workflow orchestration quality rather than one-shot model output.

\subsection{Main Interface Demonstration}

The center interface is the primary orchestration view. It uses one unified research prompt with explicit goals, constraints, and acceptance criteria. The user acts as research lead (goal confirmation and key decisions), while \VL{} handles decomposition, dispatch, and state feedback. We monitor four state objects throughout the process: \texttt{Task Graph}, \texttt{Artifact Store}, \texttt{Decision Log}, and \texttt{Execution Trace}.

Figure~\ref{fig:demo-planning} summarizes the resulting cross-view loop, in which discovered capabilities flow into goal/intent processing and explicit human decisions and are finally reflected as execution progress and synchronized task cards.

\subsection{Skills Interface Demonstration}

The left panel validates capability management during workflow execution, corresponding to annotations (1) and (2) in Figure~\ref{fig:demo-planning}. This sub-scenario contains three interactions:
\begin{itemize}
\item \textbf{Skills board browsing}: users browse available skills by research stage to quickly locate suitable capabilities.
\item \textbf{Tag-based filtering}: users filter skills by theme (e.g., Ideation, Experiment, Publication) to shorten retrieval paths.
\item \textbf{Manual skill addition}: users add new skills into the current project so they can be explicitly invoked in subsequent tasks.
\end{itemize}

This sub-scenario tests not skill count but whether users can perform \emph{Capability Discovery \& Customization} and \emph{Tag-based Filtering} in context, then convert selected skills into executable steps.

\subsection{Task List Interface Demonstration}

The right panel (\texttt{Task List}) is the execution-control view, corresponding to annotations (5) and (6) in Figure~\ref{fig:demo-planning}. It presents synchronized task status---overall counts (\texttt{Total}, \texttt{Done}, \texttt{In Progress}, \texttt{Pending}), a progress bar, and stage-grouped \emph{Synchronized Task Cards}---and supports three operations during execution: \textbf{(1) progress inspection} (assess stage completion and backlog); \textbf{(2) task-level traceability} (each card exposes task ID, objective, and linked skill tags); and \textbf{(3) direct action entry} (trigger the next step from pending items).

\section{Evaluation}

We evaluate \VL{} against the \emph{bare command-line coding agent it wraps}. The question is not whether the wrapper runs faster, since an orchestration layer that records state necessarily does more work, but whether, at a bounded time cost, it leaves the delivered output no less complete while turning a flat pile of files into an auditable, recoverable trail. The operator-facing context-switch reduction the paper claims is measured by the retrospective three-condition human study (Appendix~\ref{app:human-study}), not by this automated comparison.

\subsection{Research Completeness Under Open-Ended Goals}
\label{sec:crossover}
A fully enumerated instruction leaves little room for orchestration to add value: when every requirement
is spelled out in the prompt, a capable backend executor simply reads them off. We therefore evaluate
\VL{} against the bare command-line agent it wraps in the regime where a research assistant should
matter---an \emph{open-ended} goal, where best practices must be supplied rather than transcribed. We
hold the backend executor fixed (the \texttt{codex} provider with model \texttt{gpt-5.4} under a matched
\texttt{danger-full-access}/approval-never profile), so bare \texttt{codex} and \texttt{drclaw} differ
only by \VL{}'s task graph, artifact store, decision log, execution trace, and skill library. These
arrive as one bundle: the comparison measures what the orchestration layer adds \emph{as a whole}, and
cannot attribute the difference to any single component of it. Each
task gives an identical, unenumerated
instruction (``conduct a rigorous, publication-quality study\ldots'') across three medical problems:
melanoma and nevus classification on Derm7pt \citep{kawahara2019derm7pt}, and a clinical-note risk baseline. Neither prompt is told
the rubric. \emph{Completeness} is the fraction of \textbf{21} research best-practice elements
spontaneously included---code and reproducibility, multiple models, cross-validation, calibration,
ablation, statistical rigor, figures, a write-up citing its own numbers, plus research-hygiene elements
(a limitations section, subgroup analysis, real related-work citations)---scored deterministically against
the produced files, with no model-in-the-loop judgment. In the \VL{} condition the task graph was
verified active (mean \oeGraphTasks{} tracked tasks; the executor read $\sim$\oeSkillReads{} skill files
per run).

\begin{figure}[t]
  \centering
  \includegraphics[width=\columnwidth]{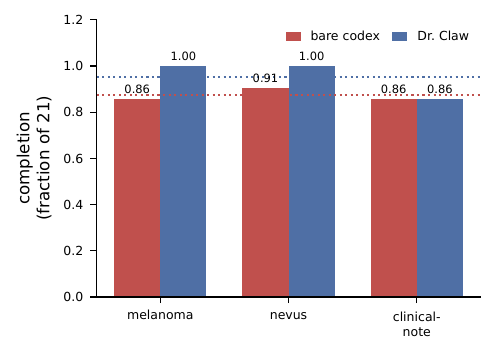}
  \vspace{-2mm}
  \vspace{-2mm}
  \vspace{-2mm}
  \vspace{-2mm}
  \vspace{-2mm}
  \caption{Open-ended completion per task (fraction of 21 elements); dotted lines are pooled means.}
  \label{fig:completion}
  \vspace{-2mm}
  \vspace{-2mm}
\end{figure}

\begin{figure}[t]
  \centering
  \includegraphics[width=\columnwidth]{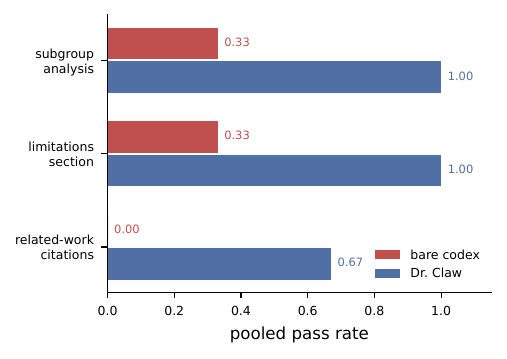}
  \vspace{-2mm}
  \vspace{-2mm}
  \vspace{-2mm}
  \vspace{-2mm}
  \vspace{-2mm}
  \caption{Pooled pass rate on the three research-hygiene elements where the conditions diverge.}
  \vspace{-2mm}
  \vspace{-2mm}
  \label{fig:hygiene}
\end{figure}

\paragraph{What the open-ended test shows.} \looseness=-1 \VL{} wins two of three tasks and ties the third, pooling
\oeDrPool{} against the bare agent's \oeBarePool{} (Figure~\ref{fig:completion}). The advantage is
\emph{not} in modeling: both conditions train three-plus models with cross-validation, calibration,
ablation, and statistical rigor---those elements pass at 1.00 on both sides. It lives entirely in
research hygiene (Figure~\ref{fig:hygiene}): \VL{}'s reference-audit, analysis, and paper-writing
skills reliably add a limitations section (\oeLimitsBare{}$\to$\oeLimitsDr{} pooled pass rate), subgroup
analysis (\oeSubgroupBare{}$\to$\oeSubgroupDr{}), and real literature citations
(\oeRelworkBare{}$\to$\oeRelworkDr{}, the bare agent producing \emph{zero} across the three tasks). On the
one tie (the clinical-note task), \VL{}'s reference audit did not fire, so it too missed citations---the
mechanism, when engaged, is exactly what closes the gap.

\paragraph{Triggering reliability.} Skill \emph{selection} is deterministic given a task's stage and type,
but skill \emph{invocation} is not enforced: the resolver can only place a suggestion in the task prompt.
Across the three runs the executor read $\sim$\oeSkillReads{} skill files per run, yet the one non-firing
reference audit above accounts for the single task on which \VL{} failed to beat the bare agent. Suggestion
is guaranteed; invocation is best-effort, and closing that gap---by verifying skill execution rather than
recommending it---is the clearest reliability improvement the current design admits.

\begin{figure}[t]
  \centering
  \includegraphics[width=\columnwidth]{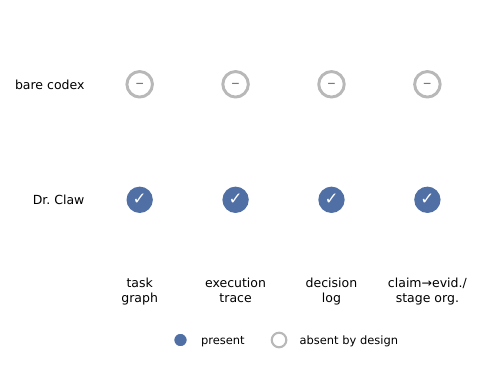}
  \vspace{-2mm}
  \vspace{-2mm}
  \vspace{-2mm}
  \vspace{-2mm}
  \vspace{-2mm}
  \caption{Whether a run persists a queryable process model (present/absent by design, not scored).}
  \label{fig:affordance}
  \vspace{-2mm}
  \vspace{-2mm}
\end{figure}

\paragraph{Auditability is an architectural affordance, not a score.} The conditions also differ in
whether a completed run can be \emph{re-traced} (Figure~\ref{fig:affordance}). Every \VL{} run persists a
\textbf{queryable task graph} (mean \oeAuTaskGraph{} nodes), a timestamped execution trace (mean
\oeAuExecTrace{} transitions), a decision-log brief (mean \oeAuDecLog{} entries), and named research
stages with explicit claim$\to$evidence maps; \textbf{the bare agent persists none}. We read this as a
\textbf{design affordance for human oversight, not a performance score}: these objects are \VL{}'s own
file format, so ``bare = absent'' holds by construction. On \emph{format-neutral} traceability
\textbf{we claim no superiority}---write-up file references resolve at \oeAuXrefDrPct{} versus
\oeAuXrefBarePct{}, but this is volume-confounded ($48$ vs $8$ references) and non-decisive at this $n$.
Both results are an exploratory pilot: with one replicate per task the pooled completion
$\Delta=\oeDeltaPool$ has a 95\% bootstrap CI of \oeCIPool{} that includes zero, so the direction is
consistent (\VL{} $\ge$ bare on all three tasks) but not yet significant, and \VL{} remains slower---a
bounded overhead the orchestration layer incurs by recording state.

\subsection{Non-Destructive Failure Recovery Under Audit}
\label{sec:casestudy}
\label{sec:recovery}

We run one coherent Derm7pt mini-project through \VL{} and recover it from an induced failure, on the same backend executor (\texttt{codex}/\texttt{gpt-5.4}). The advantage on display is not speed or accuracy but the auditable, structured artifact trail the orchestration layer maintains: when a step fails, the captured execution trace and preserved prior state let the project recover in place rather than restart.

\begin{figure}[t]
  \centering
  \includegraphics[width=1\columnwidth]{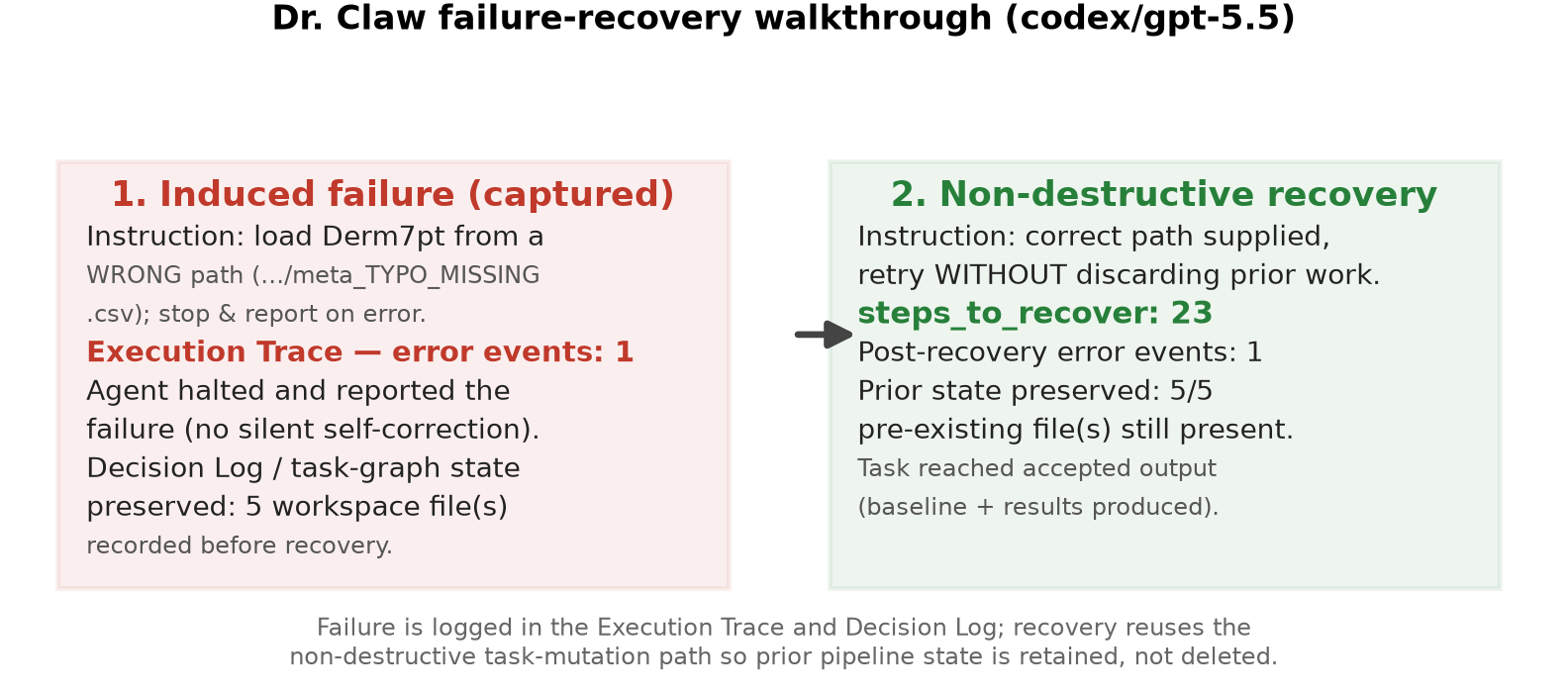}
  \vspace{-2mm}
  \vspace{-2mm}
  \vspace{-2mm}
  \caption{Captured failure and non-destructive recovery inside a project. Left: the induced wrong-path error, captured in the execution trace. Right: the in-place fix and recovered result, with all prior files retained.}
  \label{fig:recovery}
  \vspace{-2mm}
  \vspace{-2mm}
\end{figure}

\paragraph{Failure and recovery.} \looseness=-1 The same project also shows what happens when a step fails (Figure~\ref{fig:recovery}). Halting on an induced wrong-path error rather than silently self-correcting, then recovering to a real result (accuracy $0.892$) with all $5$ pre-existing files retained ($23$ tool events), the workspace \emph{adds} the corrected artifacts rather than wiping the failed attempt---the prior pipeline state survives the fix. We report this as a single-condition design demonstration, without a matched bare-agent recovery run.

\subsection{Human Study}

To measure the operator-facing effect that the automated comparison cannot, we retain a retrospective
study of seven AI PhD researchers across three research stages (Ideation, Experiment, Publication),
comparing \VL{} against working with no AI tools and with general-purpose web/desktop AI assistants
(e.g., ChatGPT, Gemini, Claude) on completion time, blind-rated output quality, tool-switching count,
and self-reported experience. Under this hybrid, exploratory protocol, \VL{} is associated with \textbf{shorter
completion-time bands, the highest output-quality ratings, and fewer tool switches with higher
experience scores}; the effect is \textbf{strongest and fully pairwise-significant for experience}. Full setup,
figures, and stage-wise statistics are in Appendix~\ref{app:human-study}.

\section{Conclusion}

We presented \VL{}, an integrated system for end-to-end AI research that unifies state-object management and skill-based execution in one workspace to reduce cross-tool orchestration costs and improve workflow continuity. Evaluated with the same backend executor run inside versus outside \VL{}, which compares the whole orchestration layer against the agent it wraps rather than ablating its parts, the layer preserves the measured completeness of the output (a count of which research components are present, not a correctness check) while producing a more auditable, better-structured artifact trail, as shown through a persisted-process-model analysis and a non-destructive failure-recovery walkthrough; a retrospective human study over three stages provides complementary evidence on time, output quality, and integrated experience.

\section*{Limitations}

Our evaluation is a small-scale, exploratory demonstration rather than a powered comparative study: the pilot uses a limited number of tasks and participants, so the reported differences are directional evidence about workflow orchestration, not causal or statistically powered effects. Three limits deserve to be stated plainly. First, holding the backend executor fixed is not an ablation: \VL{} adds a task graph, persistent state objects, a skill library, and workflow instructions as one bundle, so the observed gap cannot be attributed to any single component, and a skill-only versus orchestration-only ablation remains the natural next experiment. Second, completeness counts how many of $21$ expected research components a run produces; it is a coverage measure, not a correctness check, and establishing scientific soundness would require expert per-artifact review. Third, our comparison target is the bare executor that \VL{} wraps: the matched control for what the wrapper adds, but not a state-of-the-art orchestration framework. Other confounds remain, such as prior familiarity with either interface, and the operator-facing context-switch and intervention reductions are measured separately in the human study (Appendix~\ref{app:human-study}). Our tasks also come from a single domain (medical), so transfer of the skill library and the structured Task~Graph abstraction to other research areas, particularly open-ended work where rigid structure may add friction, remains to be shown. We do not claim model-level innovation; our contribution is workflow integration, and observed gains depend on configuration choices (backend model, permission settings, and skill coverage).

\section*{Acknowledgments}

This work was partially supported by the National Science Foundation Grants CRII-2246067, ATD-2427915, NSF POSE-2346158, and NSF POSE-2449280.

\section*{Ethics Statement}

\VL{} is designed to \emph{assist} research under sustained human control, not to autonomize it. A recurring concern with AI research systems is that they may flood the literature with unverified or low-quality output. Our design responds to this concern directly rather than amplifying it: every stage passes through explicit human checkpoints, final acceptance rests with the researcher, and the Decision~Log and Execution~Trace keep a complete, auditable record of what was generated, approved, revised, or rejected. We view this human-in-the-loop, fully-traceable structure as a safeguard for verifiability, not a shortcut around it. Critical content (citations, experimental conclusions, and manuscript claims) requires human verification before use. When sensitive data are involved, users should follow least-privilege permission settings and retain operation traces; in our own study, the medical datasets remain on the authors' server and are not redistributed, and no patient-level data are released. For high-risk domains such as healthcare, system outputs must not be used directly for real-world clinical decisions. Human participation in the user study was voluntary and based on informed consent. We used AI-based coding assistants as part of the system under study and for writing assistance, consistent with venue policy.

\bibliography{custom}

\clearpage

\appendix

\section{Human-Study Evaluation (Retrospective Three-Condition Study)}
\label{app:human-study}
\label{app:eval-protocol}

As complementary evidence to the automated evaluation in Section~5, we retain the retrospective three-condition human study from the prior submission. It compares no AI tools (\textbf{No-AI}), general-purpose web/desktop AI assistants (e.g., ChatGPT, Gemini, Claude; \textbf{Web/Desktop-AI}), and \textbf{\VL{}} over three research stages (Ideation, Experiment, Publication), on four metrics: completion-time bands, stage output score (blind rating, 1--5), switching-count bands, and experience score. The study uses a hybrid design---live logs for \VL{} and retrospective reports for the two controls---so all statistics are exploratory rather than causal.

\begin{figure*}[t]
  \centering
  \begin{subfigure}[t]{0.30\textwidth}
    \centering
    \includegraphics[width=\linewidth]{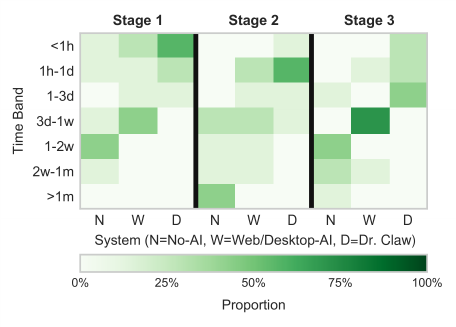}
    \caption{Time-band distributions (h/d/w/m: hour/day/week/month).}
    \label{fig:eval-time-heatmap}
  \end{subfigure}
  \hfill
  \begin{subfigure}[t]{0.30\textwidth}
    \centering
    \includegraphics[width=\linewidth]{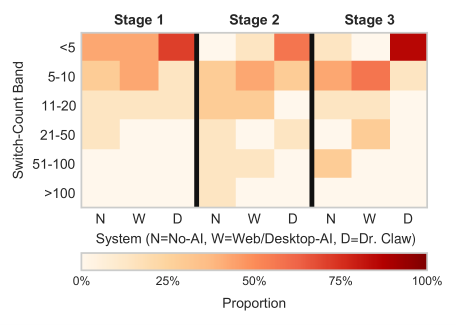}
    \caption{Switching-count band distributions.}
    \label{fig:eval-switch-heatmap}
  \end{subfigure}
  \hfill
  \begin{subfigure}[t]{0.30\textwidth}
    \centering
    \includegraphics[width=\linewidth]{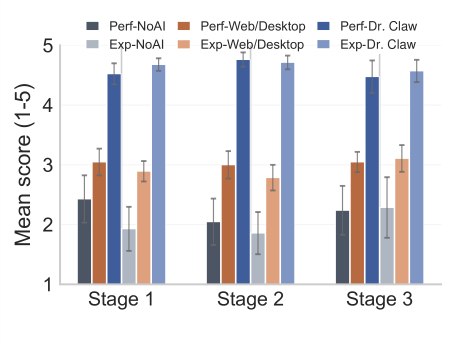}
    \caption{Mean performance (Perf) and experience (Exp); error bars: SE.}
    \label{fig:eval-mean-combined}
  \end{subfigure}
  \caption{Human-study results across three systems and three stages. Heatmap intensity denotes percentage within each system-stage cell.}
  \label{fig:eval-combined}
  \end{figure*}

\paragraph{Results.} Across all three stages, \VL{} is associated with shorter completion-time bands (Figure~\ref{fig:eval-time-heatmap}; mainly $<$1h and 1h--1d), the highest stage output scores (Figure~\ref{fig:eval-mean-combined}; \VL{}~$>$~Web/Desktop-AI~$>$~No-AI), and lower switching bands with higher experience scores (Figure~\ref{fig:eval-switch-heatmap}). The omnibus signal is strongest and fully pairwise-significant for Experience, with Time Band also significant in Stages 2--3.

\subsection{Participants and Data Collection}

We recruited seven AI PhD participants (subfields: high-performance AI, medical AI, and large language models), all with stable research and paper-writing experience and informed consent. To reduce unfamiliarity bias, each was given a three-week free-use period for \VL{} before formal comparison. Under the hybrid design, C3~(\VL{}) participants performed tasks under a unified task framework with live logs, while C1~(No-AI) and C2~(Web/Desktop-AI) were reported retrospectively (time, switching-count, and experience ranges) using the same stage definitions. Stage outputs in all conditions were scored by blind raters with the same stage-specific rubrics; to reduce recall bias, we required recallable tasks within the most recent month and a unified timing protocol.

\subsection{Stage Definitions and Rubrics}

We divide the workflow into three research stages: \textbf{Stage 1 (Ideation)} idea generation and problem framing; \textbf{Stage 2 (Experiment)} experiment setup, data processing, and analysis (\textbf{excluding model runtime}); and \textbf{Stage 3 (Publication)} drafting and final polishing. Stage output is rated on a 1--5 scale by blind raters using stage-specific rubrics: Stage~1 on novelty, feasibility, literature coverage, and clarity of problem definition; Stage~2 on reasonableness of setup, correctness of analysis, and quality of interpretation; Stage~3 on structural completeness, technical accuracy, readability, and reproducibility information.

\paragraph{Logging and statistics.} Logs are organized at participant--system--stage granularity: \VL{} live logs record start/end time, stage duration, and switching count, while retrospective questionnaire values for the controls are normalized to the same metric definitions. Because retrospective data are included, statistics are exploratory: we apply Friedman tests for overall comparisons and Holm-corrected pairwise Wilcoxon tests post-hoc.

\subsection{Detailed Statistical Results}
Stage-wise Friedman tests use $n=7$ matched participants; each entry reports $\chi^2(2), p$ per stage (S1/S2/S3), followed by Holm-corrected pairwise $p$ versus \VL{} for No-AI and Web/Desktop-AI (each as \{S1,S2,S3\}).
\begin{itemize}\itemsep2pt
\item \textbf{Time Band}: $8.96,0.0114$ / $13.56,0.0011$ / $14.00,{<}0.001$. Pairwise: No-AI \{0.1250,0.0469,0.0469\}, Web/Desktop-AI \{0.1250,0.0469,0.0469\}.
\item \textbf{Switching Band}: $1.50,0.4724$ / $11.57,0.0031$ / $10.75,0.0046$. Pairwise: No-AI \{1.0000,0.0938,0.0625\}, Web/Desktop-AI \{1.0000,0.0938,0.0469\}.
\item \textbf{Performance}: $9.25,0.0098$ / $11.57,0.0031$ / $11.31,0.0035$. Pairwise: No-AI \{0.0938,0.0938,0.0469\}, Web/Desktop-AI \{0.1250,0.0938,0.0625\}.
\item \textbf{Experience}: $12.29,0.0021$ / $13.56,0.0011$ / $12.07,0.0024$. Pairwise: both controls \{0.0469,0.0469,0.0469\}.
\end{itemize}
Key validity threats: limited sample size, system-familiarity differences, recall bias in retrospective controls, exclusion of model runtime in Stage 2, and subjectivity in experience scores.

\section{System Overview Details}
\label{app:system-details}

\subsection{Formal Model}

The write-back step is $\mathrm{Artifact}_{t+1}=\mathrm{Artifact}_t \cup \Delta \mathrm{Artifact}_t$, where $\Delta \mathrm{Artifact}_t$ is the set of newly added or revised artifacts in one iteration; task-node statuses update along dependencies (pending $\rightarrow$ running $\rightarrow$ done) with full histories retained in \texttt{Execution Trace}. \VL{}'s target follows: lowering orchestration overhead while preserving output quality and controllability.

\subsection{Implementation, Recovery, and Reproducibility}
\label{app:impl-details}
\label{app:failure-recovery}
\label{app:repro-details}

\textbf{Implementation.} Project initialization creates a fixed stage-folder layout for Ideation, Experiment, and Publication, plus a persistent pipeline-state store holding configuration, the research brief, and the task list. Task nodes store normalized fields (ID, status, priority, dependencies, stage, type, required inputs, suggested skills, next-action prompt); the server resolves status aliases and selects the next task by dependency completion, reading stage-specific skill recommendations from a stage-skill map. \textbf{Skill lifecycle.} The catalogue holds $171$ skills, $87$ of them top-level entries with their own \texttt{SKILL.md}; three in-house families supply most (\texttt{aris-*}, $44$; \texttt{inno-*}, $16$; \texttt{ds-*}, $14$), the rest imported from public collections. \emph{Authoring} is file-based: a skill is a directory whose frontmatter carries \texttt{name} and \texttt{description}, commonly \texttt{version}, \texttt{license}, \texttt{allowed-tools}, and \texttt{argument-hint}, plus optional \texttt{stage}/\texttt{domain} keys feeding the dashboard tag index. \emph{Validation} parses the frontmatter, rejects a \texttt{SKILL.md} lacking a \texttt{name}, applies pre-flight schema and dependency checks, and version-stamps activated skills for replay and audit. \emph{Selection} runs by stage-map resolution, keyword auto-load, or manual invocation: the resolver unions a stage's base skills with those for the task's type and writes them to the task node, reaching $58$ skills across five stages (survey $11$, ideation $14$, experiment $18$, publication $22$, promotion $3$); $30$ top-level skills are not yet stage-mapped. \emph{Transfer} to a new domain edits one JSON map rather than code. \VL{} uses backend adapters for the Claude and Codex SDKs (with Cursor hooks) and enforces action constraints via explicit policy settings (allowed/disallowed tools, permission mode, sandbox/approval). Pipeline mutations are written to persistent state first, then broadcast over WebSocket, keeping the interface synchronized to the same source of truth.

\textbf{Failure handling and recovery} operate at three levels: \emph{pipeline/file} (missing paths, unreadable files, and JSON parse errors return explicit 4xx/5xx responses; initialization recreates pipeline-state defaults), \emph{permission} (allow/deny checks with explicit denial reasons and bounded approval timeouts), and \emph{session} (abort-supported execution with structured error events and consistent lifecycle states). Recovery relies on non-destructive task mutation APIs---update status, revise content, append tasks, continue from pending/in-progress nodes---supporting \emph{revise/retry/handoff} recovery without deleting prior states.

\textbf{Reproducibility.} Setup requires Node.js LTS (v22 recommended), a standard install-and-run command sequence, and environment configuration. Each run should archive the instance metadata and pipeline-state files together with stage artifacts from Ideation, Experiment, and Publication. A minimum replication checklist: fix the same repository revision, lockfile, and runtime versions (Node, backend SDK/model); keep the same permission profile, three-stage task definitions with Stage-2 model-runtime exclusion, and time-/switching-band discretization; preserve blinded rubrics and rater instructions; and export raw pipeline state and run-time logs as supplementary material.

\end{document}